\documentclass[sigconf,authorversion]{acmart}

\def\methodName{\textsc{CADET}}
\usepackage{enumitem}
\usepackage{pifont}
\usepackage{subfig}   
\usepackage{xcolor} 
\usepackage{soul}   
\usepackage{listings}
\usepackage{multirow}
\usepackage{fontawesome}
\usepackage[most]{tcolorbox}
\usepackage{colortbl} 
\definecolor{RowGray}{gray}{0.95}
\usepackage{subfig}   

\theoremstyle{plain}

\AtBeginDocument{%
  }

\copyrightyear{2026}
\acmYear{2026}
\setcopyright{cc}
\setcctype{by}
\acmConference[WWW '26]{Proceedings of the ACM Web Conference 2026}{April 13--17, 2026}{Dubai, United Arab Emirates}
\acmBooktitle{Proceedings of the ACM Web Conference 2026 (WWW '26), April 13--17, 2026, Dubai, United Arab Emirates}
\acmPrice{}
\acmDOI{10.1145/3774904.3792669}
\acmISBN{979-8-4007-2307-0/2026/04}

\begin{document}

\title{Causality Guided Representation Learning for Cross-Style Hate Speech Detection}



\author{Chengshuai Zhao}
\authornote{Both authors contributed equally to this research.}
\affiliation{%
  \institution{School of Computing and Augmented Intelligence, Arizona State University}
  \city{Tempe}
  \state{AZ}
  \country{USA}
}
\email{czhao93@asu.edu}
\orcid{0000-0001-5923-626X}

\author{Shu Wan}
\authornotemark[1]
\affiliation{%
  \institution{School of Computing and Augmented Intelligence, Arizona State University}
  \city{Tempe}
  \state{AZ}
  \country{USA}
}
\email{swan@asu.edu}
\orcid{0000-0003-0725-3644}

\author{Paras Sheth}
\affiliation{%
  \institution{School of Computing and Augmented Intelligence, Arizona State University}
  \city{Tempe}
  \state{AZ}
  \country{USA}
}
\email{psheth5@asu.edu}
\orcid{0000-0002-6186-6946}

\author{Karan Patwa}
\affiliation{%
  \institution{School of Computing and Augmented Intelligence, Arizona State University}
  \city{Tempe}
  \state{AZ}
  \country{USA}
}
\email{kpatwa@asu.edu}
\orcid{0009-0007-4605-9613}

\author{K. Sel\c{c}uk Candan}
\affiliation{%
  \institution{School of Computing and Augmented Intelligence, Arizona State University}
  \city{Tempe}
  \state{AZ}
  \country{USA}
}
\email{candan@asu.edu}
\orcid{0000-0003-4977-6646}

\author{Huan Liu}
\affiliation{%
  \institution{School of Computing and Augmented Intelligence, Arizona State University}
  \city{Tempe}
  \state{AZ}
  \country{USA}
}
\email{huanliu@asu.edu}
\orcid{0000-0002-3264-7904}

\renewcommand{\shortauthors}{Chengshuai Zhao et al.}

\begin{abstract}
The proliferation of online hate speech poses a significant threat to the harmony of the web. While explicit hate is easily recognized through overt slurs, implicit hate speech is often conveyed through sarcasm, irony, stereotypes, or coded language---making it harder to detect. Existing hate speech detection models, which predominantly rely on surface-level linguistic cues, fail to generalize effectively across diverse stylistic variations. Moreover, hate speech spread on different platforms often targets distinct groups and adopts unique styles, potentially inducing spurious correlations between them and labels, further challenging current detection approaches. Motivated by these observations, we hypothesize that the generation of hate speech can be modeled as a causal graph involving key factors: contextual environment, creator motivation, target, and style. Guided by this graph, we propose \methodName{}, a causal representation learning framework that disentangles hate speech into interpretable latent factors and then controls confounders, thereby isolating genuine hate intent from superficial linguistic cues. Furthermore, \methodName{} allows counterfactual reasoning by intervening on style within the latent space, naturally guiding the model to robustly identify hate speech in varying forms. \methodName{} demonstrates superior performance in comprehensive experiments, highlighting the potential of causal priors in advancing generalizable hate speech detection.\footnote{Code:~\href{https://github.com/Shu-Wan/cadet}{https://github.com/Shu-Wan/cadet}}

\end{abstract}




\maketitle

\begin{figure}[!th]
    \centering
    \includegraphics[width=1\linewidth]{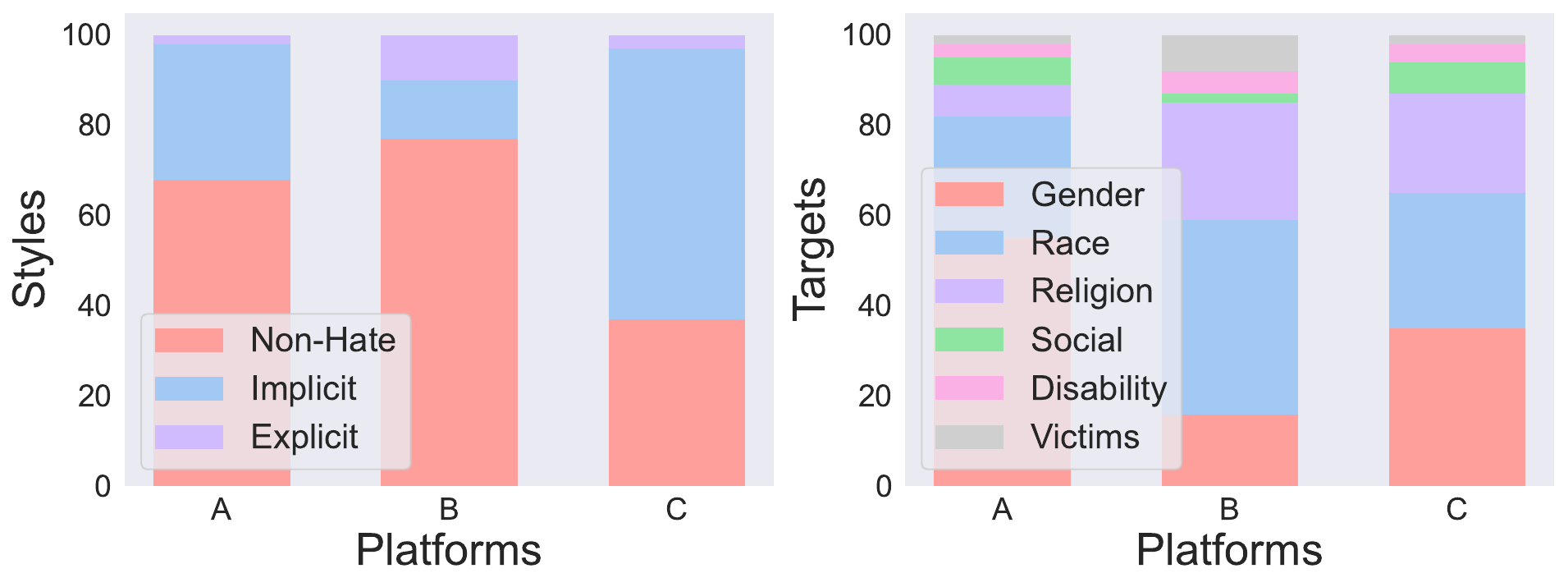}
    \caption{Distribution of hate speech styles and target groups across various social media platforms.}
    \label{fig:illustration}
    \vspace{-2mm}
\end{figure}

\section{Introduction}
The proliferation of online hate speech poses a significant threat to the harmony of the web due to its complex and ever-evolving nature~\cite{castano2021internet,malmasi2017detecting,mathew2019spread}. While considerable progress has been made in identifying explicit hate speech characterized by overt hostility, slurs, and direct attacks, the detection of implicit expressions of hate conveyed through sarcasm, irony, stereotype, or coded language remains a critical unresolved challenge~\cite{schmidt2017survey}.

Consider the examples: (i) ``\textit{F****** women asking for equal rights here. Why don't you go whine and w**** around you floozy sl**.}'' and (ii) ``\textit{I am so tired of doing everything on my own.. Need a woman so that chores can be taken care of...}''. Both express misogynistic sentiments, but while the explicit version uses direct derogatory language that is easily flagged by conventional detection systems, the implicit version conveys the same prejudice through innuendo and rhetorical framing that is much harder to detect algorithmically.

Further analysis of current hate speech detection systems reveals a persistent pattern: models trained predominantly on explicit hate speech data perform poorly when confronted with implicit expressions~\cite{ocampo2023unmasking,waseem2017understanding}. This limitation arises because these models often rely heavily on surface-level patterns such as specific slurs, aggressive syntax, or overtly hostile language that are typically absent in implicit hate speech, rather than the underlying hate intent that is shared across hate speech in various styles. Moreover, hate speech disseminated on various platforms often targets distinct groups and exhibits unique stylistic variations, as shown in Figure~\ref{fig:illustration}. This can induce spurious correlations between these factors and hate labels, posing additional challenges for existing detection models.

Motivated by these observations, we \textit{\textbf{hypothesize}} that the generation of hate speech can be modeled as a causal graph involving key factors: contextual environment (e.g., platform), creator motivation (e.g., hate intent), target, and style, where (i) contextual environment is a confounder, which influences the motivation, target, and style. (ii) Motivation, target, and style jointly shape the content of an online post. (iii) The hate label is determined by the creator's motivation and contextual environment. Guided by this graph, we propose \methodName{}, a causal representation learning framework for cross-style hate speech detection. \methodName{} first projects online posts into a latent space and disentangles them into causality-aligned interpretable factors, isolating genuine hate intent from superficial linguistic cues. Then it leverages a confounder mitigation module to control the confounding effect caused by the contextual environment. Furthermore, \methodName{} allows counterfactual reasoning by intervening on style within the latent space, naturally guiding the model to robustly identify hate speech in varying forms.

Our comprehensive evaluation demonstrates \methodName{} achieves superior performance across multiple challenging scenarios. In cross-style generalization tasks, \methodName{} achieves 0.81 average macro-F1, acquiring a relative improvement of 13\% compared with the state-of-the-art method. Ablation studies confirm the critical role of each component in \methodName{}. Further latent representation visualization and further analysis demonstrate the effectiveness of causal-driven design. Our work underscores the potential of causal disentanglement in improving generalized hate speech detection, offering practical implications for creating safer and more responsible online environments. Our contributions are summarized:
\begin{itemize}[leftmargin=*]
    \renewcommand{\labelitemi}{$\star$}
   \item We introduce a principled causal graph that explicitly models the generative process of hate speech, systematically disentangling online posts into interpretable latent factors. It unifies explicit and implicit hate speech within a coherent causal framework.
   \item We design and implement \methodName{}, a representation learning framework that operationalizes the proposed causal graph. By isolating latent factors and leveraging counterfactual reasoning, \methodName{} effectively mitigates platform-induced confounding and robustly generalizes across expression styles.
   \item We conduct extensive experiments on multiple real-world hate speech datasets, demonstrating that \methodName{} significantly outperforms state-of-the-art baselines in cross-style hate speech detection, providing robust, interpretable, and causally informed insights into harmful online content moderation.
\end{itemize}

\section{Related Work}
\subsection{Hate Speech Detection}
Early detection systems rely on lexicon-based filters~\cite{tanev2024jrc, gitari2015lexicon}. While precise for overt insults, these methods suffer from low recall, failing to capture sarcasm, metaphor, or coded language. The advent of labeled corpora shifts the focus to supervised classification; early approaches utilize logistic regression and SVMs with n-gram or TF–IDF features~\cite{talat2016hateful, davidson2017automated}. However, these models often overfit to dataset-specific artifacts, limiting cross-domain generalization~\cite{schmidt2017survey, fortuna2018survey}. The transition to deep learning--spanning CNNs, LSTMs, and transformers--enables the capture of richer semantic context~\cite{malik2024deep, caselli2021hatebert, mathew2021hatexplain}. Domain-adaptive pre-training (e.g., HateBERT~\cite{caselli2021hatebert}) and explainable architectures (e.g., HateXplain~\cite{mathew2021hatexplain}) further advance the state-of-the-art. Despite these gains, models remain brittle against implicit hate. Recent datasets on latent hatred~\cite{elsherief2021latent}, coded terms~\cite{saleh2023detection}, and adversarial shifts~\cite{rottger2021hatecheck, vidgen2020learning} highlight significant gaps in handling subtle or context-dependent expression. \methodName{} addresses this by prioritizing underlying intent and context over surface-level linguistics, enabling robust detection of implicit hate.

\subsection{Domain Generalization in Hate Speech Detection}
Hate speech manifests across diverse online platforms and in varying styles, raising the challenge of domain generalization---whether models trained in one context can robustly transfer to others~\cite{corazza2019cross,phadke2020many,sheth2023peace}. Studies consistently show that hate speech classifiers trained on one dataset or platform perform poorly when deployed elsewhere, due to reliance on domain-specific artifacts~\cite{arango2019hate, zhang2023mitigating}. For instance, models fixating on particular slurs or minority identifiers may miss hate speech expressed differently on new platforms~\cite{vidgen2019challenges}. To address this, research has explored pooling datasets from multiple domains~\cite{arango2019hate}, encouraging models to learn features less tied to any single source. Domain-invariant representation learning, especially via adversarial training, is widely used: a domain discriminator encourages the model to ignore platform-specific signals~\cite{ganin2016domain}. This has been adapted for hate speech with frameworks like MultiFOLD~\cite{arango2024multifold}, which applies curriculum learning and adversarial losses to achieve cross-domain adaptation. Some works seek ``universal” cues--such as sentiment or aggression--that might generalize, but such hand-engineered features risk missing critical nuance~\cite{wiegand2019detection, sarwar2022unsupervised}. A particularly hard case is style shift: explicit versus implicit hate. The study demonstrates that transformer embeddings separate explicit hate well while clustering implicit hate with benign content, creating a style gap~\cite{ocampo2023unmasking}. Their solution links implicit and explicit expressions targeting the same group, but full cross-style transfer remains elusive. Our work disentangles style as a latent factor and leverages counterfactual reasoning to generate explicit/implicit variants of the same content, thereby enforcing style-invariant predictions.

\subsection{Causal Representation Learning for Hate Speech Detection}
Causal representation learning aims to discover latent variables that correspond to the true generative mechanisms behind data~\cite{scholkopf2021toward,ahuja2023interventional,brehmer2022weakly}. This is especially desirable in hate speech detection, where models should attend to harmful intent or prejudice rather than superficial cues tied to specific words or platforms~\cite{sheth2024causality}. Recent work has applied causal reasoning to mitigate model bias: recent efforts attempt to construct a causal graph to identify how spurious correlations (e.g., stereotype-linked co-occurrences) drive false positives, using counterfactual data augmentation and multi-task learning to weaken such biases~\cite{zhang2023mitigating}. Other approaches generate synthetic examples by altering identity, style, or target in hateful posts, encouraging invariance to non-causal factors\cite{sen2022counterfactually, tiwari2022robust, pamungkas2019cross}. A leading line of research seeks to disentangle causal factors in the representation space. The CATCH framework~\cite{sheth2024causality} models hate targets as confounding factors linked to platform distribution. By separating target-specific and invariant components, CATCH improves cross-platform transfer. However, most prior work considers only one or two latent factors, such as target or platform, which is insufficient for detecting deceptive implicit hatch speech. Our proposed \methodName{} advances this direction by proposing a fine-grained causal model with four key latent factors: creator motivation, target, style, and context. \methodName{} disentangles these factors through orthogonal and adversarial losses, ensuring independence and bridging the gap between implicit and explicit hate speech detection.

\begin{figure}[!th]
    \centering
    \includegraphics[width=0.9\linewidth]{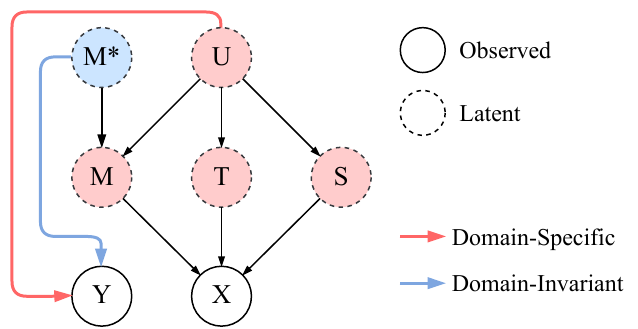}
    \caption{Causal graph of hate-speech generation. Creator motivation (M), target (T), and style (S) are influenced by contextual environment (U). M, T, and S jointly determine the post (X). The hate label (Y) depends on genuine motivation M* and U. M* is independent of unobserved confounder U.}
    \label{fig:causal_graph}
\end{figure}

\section{The Hypothesis: Hate Speech Generation}
To move beyond surface-level pattern recognition, we must separate the causal factors that render content hateful from the spurious factors that determine how such hate is conveyed.

To this end, we propose a causal graph that formalizes the generative process of hate speech as showcased in Figure~\ref{fig:causal_graph}. \textit{Post} (X), representing the textual content produced by a creator, and the \textit{Label} (Y), a binary indicator of whether the post is hateful. They are both accessible during training. The post X is generated by three latent factors: \textit{Creator motivation} (M), capturing the writer’s intent for desired impact; \textit{Target} (T), denoting the individual or group toward whom the post is directed; and \textit{Style} (S), describing the linguistic expression of the content. These latent variables and the label are influenced by a shared, unobserved \textit{Contextual environment} confounder (U). U represents platform-level factors such as moderation policies, demographics, as well as broader sociopolitical atmosphere. Differences in moderation strictness or audience tolerance can simultaneously shape a creator’s stylistic choices, target selection, and even how intent is shaped in practice, thereby inducing spurious correlations among M, T, S, and the label Y. To separate genuine intent from its context-dependent manifestation, we posit an additional latent variable M*, representing the creator’s true underlying motivation or intent, which is invariant to the contextual environment U and a direct cause of the label Y.

Unlike conventional causal formulations in domain generalization that assume domain shift affects either X or Y alone~\cite{liu2021rethinking}, our graph models domain shift as an unobserved confounder influencing both X and Y. This design reflects current online discourse, where content creators actively adapt their writing to the contextual environment U, often shifting from explicit to implicit or obfuscated expressions to evade moderation. As a result, classifiers trained predominantly on explicit-style corpora struggle to detect implicit hate in new domains. In our proposed graph, while the observed latents (M, T, S) are entangled with U, the true motivation M* is independent of the contextual environment. Hence, M* is a domain-invariant causal feature that is optimal and robust for out-of-distribution prediction \cite{wang2022causal}. Therefore, learning M* is the goal of \methodName{}. Additionally, the graph also motivates a principled path toward cross-style generalization through counterfactual reasoning. By intervening on the style (S) node while holding creator motivation (M) and target (T) fixed, we can construct counterfactual examples that express the same hateful intent in alternative stylistic forms. Training models to treat both original and counterfactual samples as hateful encourages representations that emphasize invariant causal signals rather than superficial stylistic cues.

\section{The Proposed CADET}
In this section, we formally define the problem (Section~\ref{sec:problem}) and present CADET, a causality-guided representation learning framework for cross-style hate speech detection. CADET comprises three core components: (i) a causally-aligned disentanglement architecture, which separates generative factors in the latent space according to the causal graph (Section~\ref{sec:disentanglement}); (ii) a confounder mitigation mechanism, which reduces the impact of spurious correlations and alleviates confounding effects (Section~\ref{sec:confounder}); and (iii) a counterfactual reasoning module, which promotes style-invariant representations by generating and leveraging counterfactual examples (Section~\ref{sec:counterfactual}).

\begin{figure*}[th]
    \centering
    \includegraphics[width=\textwidth]{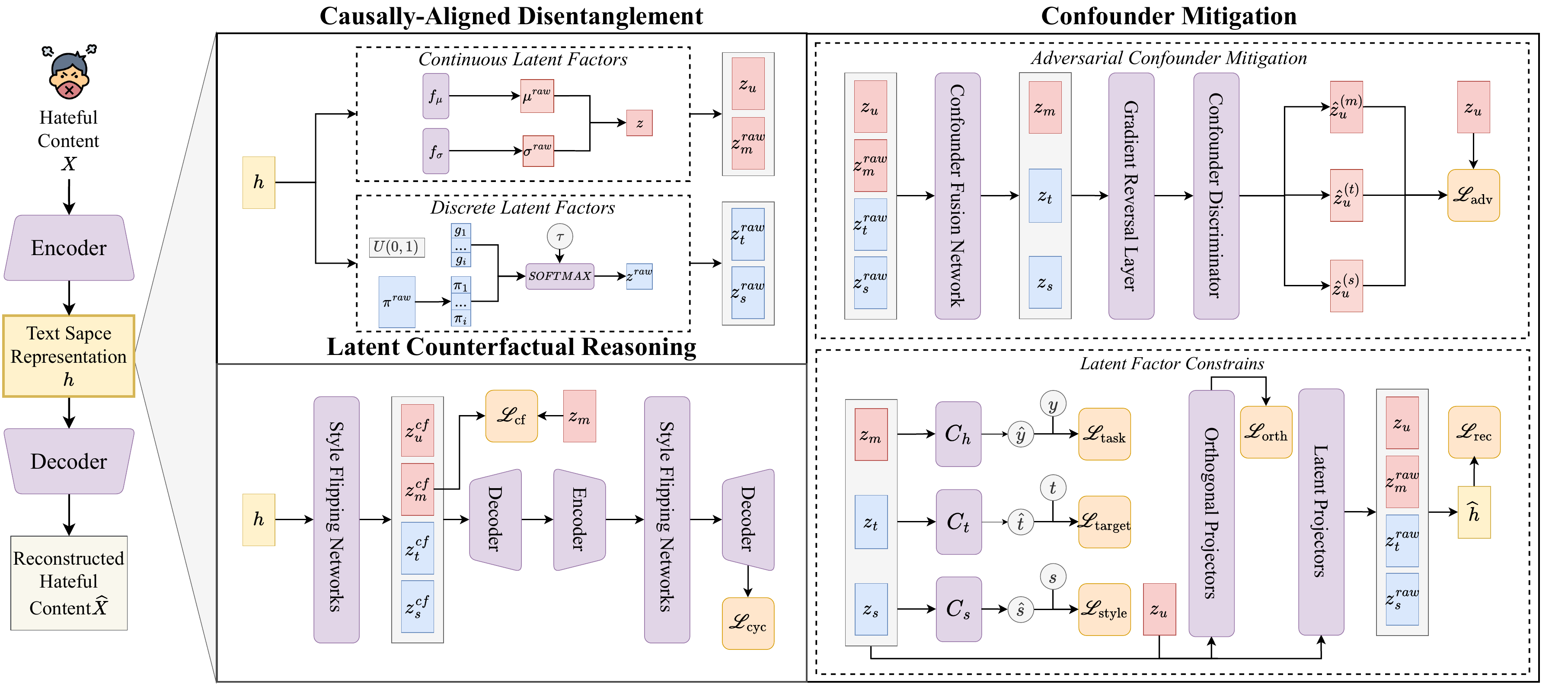}
    \caption{Framework of the proposed method. \methodName{} firstly decomposes the factor in the latent space guided by the causal graph. Then, it mitigates the confounding effect of the platform, where various constraints are applied to enable the meaning representation learning. various. Finally, latent counterfactual reasoning is employed to learn style-invariant features.}
    \label{fig:cadet_architecture}
\end{figure*}

\subsection{Problem Definition}
\label{sec:problem}
Let $\mathcal{D} = \{(x_i, y_i, s_i, t_i)\}_{i=1}^n$ denote a corpus of social media posts, where $x_i$ denotes the textual content, $y_i \in \{0,1\}$ is a binary hate speech label, $s_i \in \{0,1\}$ indicates whether the post conveys hate speech explicitly or implicitly, and $t_i$ (optional) specifies the targeted group. We define $\mathcal{D}_{s}$ as the conditional distribution over $(x,y)$ given a particular style $s$. We formalize \emph{cross-style hate speech detection} as a domain generalization problem: learning an optimal predictor $f^*$ that is trained on posts from a source style domain $\mathcal{D}_{s}$ and generalizes effectively to a target style domain $\mathcal{D}_{s'}$ with $s \neq s'$.
\begin{equation}
f^* = \arg\min_{f \in \mathcal{F}} \;
\mathbb{E}_{(x,y) \sim \mathcal{D}_{s'}} \big[ \ell(f(x), y) \big],
\end{equation}
where $\ell$ denotes the loss function and $\mathcal{F}$ is the hypothesis space.

\subsection{Causally-Aligned Disentanglement}
\label{sec:disentanglement}
To enable the causality-guided representation disengagement, we first map online posts into a text latent space.

\subsubsection{Text Encoding}
We utilize a pretrained RoBERTa model as an encoder $E$ to transform input text $x_i$ into latent representations $h$.
\begin{equation}
h = \text{Encoder}(x)
\end{equation}
where $h \in \mathbb{R}^d$ is the embedding of the start-of-sequence token (CLS) that serves as a fixed-length representation of the entire post.

\subsubsection{Latent Factor Inference.}
We infer four latent variables from $h$ that correspond to the causal factors in our graph: we model the confounder representation $z_u$ as a continuous Gaussian variable, parameterized via the reparameterization trick:
\begin{equation}
z_u = \mu_u + \sigma_u \odot \epsilon, \quad \epsilon \sim \mathcal{N}(0,I),
\end{equation}
where the mean and variance parameters are predicted by feed-forward networks, $\mu_u = F_{\mu_u}(h)$ and $\log \sigma_u^2 = F_{\log \sigma_u^2}(h)$. Similarly, we extract the creator motivation representation, denoted $z_m^{\text{raw}}$, which is also represented as a continuous Gaussian variable.

The target factor $z_t^{\text{raw}}$ is modeled as a discrete variable, parameterized using the Gumbel–Softmax trick:
\begin{equation}
z_t^{\text{raw}} = \text{GumbelSoftmax}(\pi_t^{\text{raw}}, \tau_t),
\end{equation}
where $\pi_t^{\text{raw}} = F_t(h)$ are the logits associated with the target and $\tau_t$ is the temperature parameter controlling smoothness. Similarly, the content style, denoted $z_s^{\text{raw}}$, is represented as a discrete binary variable (capturing explicit vs. implicit style) inferred using the Gumbel–Softmax parameterization.

\subsection{Confounder Mitigation}
\label{sec:confounder}
To alleviate the impact of spurious correlations introduced by the context, we design a confounder-mitigation mechanism.

\subsubsection{Adversarial Confounder Reconstruction.} We employ confounder fusion networks $F$ to control the latent confounder by merging it into the raw variables:
\begin{equation}
    z_i = \sigma_i\bigl(F_i(z_i^{\text{raw}} || z_u)\bigr), \; i \in \{m, t,s\}
\end{equation}
where $\sigma$ is an activation denoting the identity for continuous factors (e.g., creating motivation) and the softmax function for discrete factors (e.g., target and style), and each fusion network is implemented as a feed-forward network with layer normalization. 

To ensure proper and meaningful disentanglement of each causal factor, we attach adversarial discriminators $D$ that attempt to recover the confounder from each merged variable:
\begin{equation}
    \hat{z}_u^{(i)} = D_i\bigl(G_\lambda(z_i)\bigr), \; i\in\{m,t,s\}
\end{equation}
where $G_\lambda$ is a gradient-reversal layer parameterized by the scaling coefficient $\lambda$ so that the feature extractor is updated against the adversary. The resulting adversarial loss is defined by:
\begin{equation}
\mathcal{L}_{\text{adv}} = ||\hat{z}_u^{(m)} - z_u||_2^2 + ||\hat{z}_u^{(t)} - z_u||_2^2 + ||\hat{z}_u^{(s)} - z_u||_2^2
\end{equation}

\subsubsection{Factor-Specific Objective.} We incorporate causally-aligned classification, ensuring that each classifier relies solely on its causally relevant factor. We feed fused motivation representation into classifiers $C_h$ to obtain the predicted label $\hat{y}$:
\begin{equation}
\hat{y} = C_h(z_m)
\end{equation}
This architectural constraint directly enforces the causal understanding that hate depends on the creator's motivation. Then, we utilize cross-entropy to compute the loss:
\begin{equation}
    \mathcal{L}_{\text{task}} = \text{CrossEntropy}(\hat{y},y)
\end{equation}

Similarly, we can acquire target loss $\mathcal{L}_{\text{target}}$ and style loss $\mathcal{L}_{\text{style}}$ by decoding corresponding latent factors.

\subsubsection{Orthogonality Constraints} Additionally, orthogonality constraints are used to encourage separation of latent factors. We employ projections to map each variable into a common space:
\begin{equation}
p_i = F_{\text{orth}}^{(i)}(z_i), \; i\in\{m,t,s,u\}
\end{equation}
and compute pair-wise squared dot products between latent factors:
\begin{equation}
o_{ij} = \left(\frac{p_i}{\|p_i\|_2} \cdot \frac{p_j}{\|p_j\|_2}\right)^2, \; (i,j)\in\{m,t,s,u\}^2, \; i < j
\end{equation}

In this way, we obtain the orthogonality loss:
\begin{equation}
\mathcal{L}_{\text{orth}} = o_{mt} + o_{ms} + o_{ts} + \lambda_u (o_{mu} + o_{tu} + o_{su})
\end{equation}
where $\lambda_u > 1$ places stronger regularization with the confounder.

\subsubsection{Input Reconstruction}
We use reconstruction as a constraint to ensure our latent spaces capture meaningful information. Each latent variable is projected to the decoder's hidden dimension:
\begin{equation}
h_i = F_\text{Proj}^{(i)}(z_i), \; i\in\{m,t,s,u\}.
\end{equation}

These projections are combined into a fused representation:
\begin{equation}
\hat{h} = h_m + h_t + h_s + h_u
\end{equation}

We use a pretrained BART decoder $\text{Decoder}$ to reconstruct the original input:
\begin{equation}
\hat{x} = \text{Decoder}(\tilde{x}, \hat{h})
\end{equation}
where $\tilde{x}$ represents the input sequence shifted right by one position for teacher forcing. The reconstruction loss is:
\begin{equation}
\mathcal{L}_{\text{rec}} = \text{CrossEntropy}(\hat{x}, x)
\end{equation}

\subsection{Latent Counterfactual Reasoning}
\label{sec:counterfactual}
To learn style-invariant representations, CADET leverages latent counterfactual reasoning without generating harmful text. 

\subsubsection{Style Intervention}
Given latent representation $h$, we leverage a style flipping network $R$ to disentangle latent factors by inverting the style component while maintaining other factors:
\begin{equation}
    R(z): \; z_i^{\text{cf}} =
    \left\{\begin{array}{ll}
    I - \text{OneHot}(z_s), \; i = s\\
    \text{Identical}(z_i), \; i \in \{m, t, u\}
    \end{array}\right.
\end{equation}
The binary nature of our style factors enables generalization even when the training data contains only one style.

\subsubsection{Counterfactual Consistency.}
Since hate should depend only on motivation (instead of style), the output of the classifier should remain stable under style transformations:
\begin{equation}
    \mathcal{L}_{cf} = ||C_h(z_m) - C_h(z_m^{\text{cf}})||_2^2
\end{equation}

\subsubsection{Cycle Consistency}
We further implement a cycle consistency mechanism to ensure style transformations preserve semantic content. It decodes the counterfactual latent representation. Then it attempts to leverage another style flipping network to reverse the style back. Finally, we can construct the original text.
\begin{equation}
\mathcal{L}_{\text{cycle}} = \mathcal{L}_{\text{rec}}(x, \text{Decoder}(R(\text{Encoder}(\text{Decoder}(\hat{h}^\text{cf})))))
\end{equation}

\subsection{Model Training}
We train CADET using a multi-objective loss function that combines reconstruction, classification, disentanglement, and counterfactual consistency objectives. The total loss is defined as:
\begin{equation}
\begin{split}
\mathcal{L} = \ &\lambda_{\text{task}}\mathcal{L}_{\text{task}} + \lambda_{\text{target}}\mathcal{L}_{\text{target}} + \lambda_{\text{style}}\mathcal{L}_{\text{style}} + \lambda_{\text{orth}}\mathcal{L}_{\text{orth}} + \\
&\lambda_{\text{adv}}\mathcal{L}_{\text{adv}} +
\lambda_{\text{rec}}\mathcal{L}_{\text{rec}} + \lambda_{\text{cf}}\mathcal{L}_{\text{cf}} + \lambda_{\text{cycle}}\mathcal{L}_{\text{cycle}} +  + \lambda_{\text{KL}}\mathcal{L}_{\text{KL}}
\end{split}
\end{equation}
where $\mathcal{L}_{\text{KL}}$ terms regularize the latent distributions.

This streamlined process allows CADET to detect hate speech regardless of its stylistic presentation, providing a principled solution to the challenge of style-invariant detection.

\section{Experiments}
In this section, we implement comprehensive experiments to evaluate the effectiveness of our proposed CADET model for cross-style hate speech detection. These experiments directly test the causal hypotheses introduced in our framework, examining how well representations learned generalize across different hate speech styles.

\subsection{Datasets and Evaluation}
Our experiments leverage four hate speech datasets: IsHate~\cite{ocampo2023depth}, Implicit Hate Corpus (IHC), AbuseEval v1.0~\cite{caselli2020feel}, and DynaHate~\cite{vidgen2021learning} as summarized in Table~\ref{tab:datasets}. They encompass hate speech in various styles across different platforms, providing a comprehensive evaluation. The detailed information is listed in Appendix~\ref{app:dataset}.

We evaluate cross-style hate speech detection across two settings: \textit{explicit-to-implicit} and \textit{implicit-to-explicit}. Performance is measured using precision, recall, and macro-F1, averaged over five independent runs to ensure statistical robustness. We benchmark \methodName{} against four categories of baselines: (i) \textbf{Classic NLP models:} BERT~\cite{devlin2019bert}, RoBERTa~\cite{liu2019roberta}, BART~\cite{lewis2020bart}, and DistilBERT~\cite{sanh2019distilbert}; (ii) \textbf{Traditional hate speech detectors:} HateBERT~\cite{caselli2021hatebert} and HateXplain~\cite{mathew2021hatexplain}; (iii) \textbf{Causal representation methods:} PEACE~\cite{sheth2023peace} and HateWATCH~\cite{sheth2024cross}; and (iv) \textbf{LLM-based moderation models} -- ShieldGemma 2B~\cite{zeng2024shieldgemma} and Llama Prompt Guard 2~\cite{dubey2024llama}.

\begin{table}[!ht]
\centering
\caption{Statistics of hate speech datasets.}
\resizebox{\linewidth}{!}{%
\begin{tabular}{l l c c c}
\toprule
\textbf{Dataset} & \textbf{Platform} & \textbf{\# Text} & \textbf{\% Hate Class} & \textbf{\% Implicit} \\
\midrule
IsHate              & Multiple   & 29,116  & 38.6 & 11.0 \\
IHC                 & Twitter    & 22,584  & 39.6 & 96.8 \\
AbuseEval           & Twitter    & 14,100  & 32.9 & 34.2 \\
DynaHate            & Synthetic  & 41,134  & 53.9 & 83.3 \\
\bottomrule
\end{tabular}%
}
\label{tab:datasets}
\end{table}

\begin{table*}[!th]
\small
\centering
\renewcommand{\arraystretch}{0.90}
\caption{Cross-style Generalization Performance. \textbf{Bold} indicates the best performance and \underline{underline} the second best.}
\label{tab:main}
\resizebox{\linewidth}{!}{
\begin{tabular}{lcccccccccccc}
\toprule
\textbf{Dataset~$\rightarrow$} &
\multicolumn{3}{c}{\textbf{IsHate}} &
\multicolumn{3}{c}{\textbf{IHC}} &
\multicolumn{3}{c}{\textbf{AbuseEval}} &
\multicolumn{3}{c}{\textbf{DynaHate}} \\
\cmidrule(lr){2-4} \cmidrule(lr){5-7} \cmidrule(lr){8-10} \cmidrule(lr){11-13}
  \textbf{Model.~$\downarrow$} &
  \textbf{Pre.~$\uparrow$} & \textbf{Rec.~$\uparrow$} & \textbf{F1.~$\uparrow$} &
  \textbf{Pre.~$\uparrow$} & \textbf{Rec.~$\uparrow$} & \textbf{F1.~$\uparrow$} &
  \textbf{Pre.~$\uparrow$} & \textbf{Rec.~$\uparrow$} & \textbf{F1.~$\uparrow$} &
  \textbf{Pre.~$\uparrow$} & \textbf{Rec.~$\uparrow$} & \textbf{F1.~$\uparrow$} \\
\midrule

\multicolumn{13}{c}{\textit{Explicit $\rightarrow$ Implicit}} \\
\midrule
BERT & 0.69 & \textbf{1.00} & 0.82 & 0.37 & \textbf{0.87} & 0.52 & 0.14 & \textbf{1.00} & 0.25 & 0.68 & \underline{0.87} & 0.76 \\
\rowcolor{RowGray}
RoBERTa & 0.69 & \textbf{1.00} & 0.82 & 0.38 & \underline{0.85} & 0.53 & 0.14 & \textbf{1.00} & 0.25 & 0.69 & \textbf{0.89} & \underline{0.78} \\
BART & 0.71 & \underline{0.99} & 0.82 & 0.38 & 0.81 & 0.51 & 0.14 & \textbf{1.00} & 0.25 & \underline{0.70} & 0.85 & 0.77 \\
\rowcolor{RowGray}
DistilBERT & 0.70 & \underline{0.99} & 0.82 & 0.38 & 0.80 & 0.52 & 0.14 & \textbf{1.00} & 0.25 & 0.66 & \underline{0.87} & 0.75 \\
HateBERT & 0.60 & 0.65 & 0.62 & 0.48 & 0.60 & 0.53 & 0.56 & 0.64 & 0.60 & 0.64 & 0.66 & 0.65 \\
\rowcolor{RowGray}
HateXplain & 0.58 & 0.63 & 0.60 & 0.54 & 0.60 & 0.57 & 0.57 & 0.61 & 0.59 & 0.58 & 0.60 & 0.59 \\
PEACE & 0.72 & 0.74 & 0.73 & \underline{0.70} & 0.72 & \underline{0.71} & \underline{0.61} & 0.63 & \underline{0.62} & 0.61 & 0.63 & 0.62 \\
\rowcolor{RowGray}
HateWATCH & \underline{0.95} & 0.93 & \underline{0.94} & 0.62 & 0.64 & 0.63 & 0.58 & 0.60 & 0.59 & 0.64 & 0.64 & 0.64 \\
ShieldGemma 2B & 0.88 & 0.74 & 0.80 & 0.49 & 0.79 & 0.60 & 0.25 & 0.31 & 0.28 & \underline{0.70} & 0.82 & 0.76 \\
\rowcolor{RowGray}
Llama Prompt Guard 2 & 0.43 & 0.00 & 0.00 & 0.23 & 0.00 & 0.00 & 0.13 & 0.03 & 0.05 & 0.00 & 0.00 & 0.00 \\
\textbf{CADET (Ours)} & \textbf{0.97} & 0.95 & \textbf{0.96} & \textbf{0.80} & 0.78 & \textbf{0.79} & \textbf{0.68} & \underline{0.67} & \textbf{0.67} & \textbf{0.87} & 0.81 & \textbf{0.84} \\
\midrule

\multicolumn{13}{c}{\textit{Implicit $\rightarrow$ Explicit}} \\
\midrule
BERT & \textbf{0.99} & 0.49 & 0.66 & \textbf{0.80} & 0.77 & \underline{0.79} & \underline{0.98} & 0.44 & 0.61 & \underline{0.92} & \underline{0.88} & \underline{0.90} \\
\rowcolor{RowGray}
RoBERTa & \textbf{0.99} & 0.47 & 0.64 & \underline{0.78} & 0.77 & 0.78 & \underline{0.98} & 0.48 & 0.65 & \textbf{0.93} & \underline{0.88} & \underline{0.90} \\
BART & \textbf{0.99} & 0.50 & 0.67 & \underline{0.78} & \underline{0.79} & \underline{0.79} & \underline{0.98} & 0.47 & 0.64 & \underline{0.92} & 0.86 & 0.89 \\
\rowcolor{RowGray}
DistilBERT & \textbf{0.99} & 0.47 & 0.64 & \underline{0.78} & 0.78 & 0.78 & \underline{0.98} & 0.44 & 0.61 & 0.91 & 0.82 & 0.86 \\
HateBERT & 0.72 & 0.70 & 0.71 & 0.67 & 0.64 & 0.65 & 0.48 & 0.46 & 0.47 & 0.52 & 0.50 & 0.51 \\
\rowcolor{RowGray}
HateXplain & 0.68 & 0.66 & 0.67 & 0.63 & 0.61 & 0.62 & 0.44 & 0.42 & 0.43 & 0.65 & 0.63 & 0.64 \\
PEACE & 0.72 & 0.70 & 0.71 & 0.67 & 0.65 & 0.66 & 0.74 & \underline{0.72} & \underline{0.73} & 0.54 & 0.52 & 0.53 \\
\rowcolor{RowGray}
HateWATCH & 0.91 & \underline{0.89} & \underline{0.90} & 0.61 & 0.59 & 0.60 & 0.71 & 0.69 & 0.70 & 0.72 & 0.70 & 0.71 \\
ShieldGemma 2B & \underline{0.94} & 0.86 & \underline{0.90} & 0.49 & \underline{0.79} & 0.60 & 0.25 & 0.31 & 0.28 & 0.70 & 0.82 & 0.76 \\
\rowcolor{RowGray}
Llama Prompt Guard 2 & 0.50 & 0.00 & 0.00 & 0.23 & 0.00 & 0.00 & 0.12 & 0.03 & 0.05 & 0.00 & 0.00 & 0.00 \\
\textbf{CADET (Ours)} & 0.93 & \textbf{0.91} & \textbf{0.92} & 0.74 & \textbf{0.89} & \textbf{0.81} & \textbf{0.98} & \textbf{1.00} & \textbf{0.99} & 0.88 & \textbf{0.93} & \textbf{0.91} \\
\bottomrule
\end{tabular}
}
\end{table*}

\subsection{Implementation Details}
\label{sec:implementation}
Training requires carefully balancing multiple loss components. We found that a straightforward joint optimization of all objectives was challenging. Therefore, we leverage a staged loss-weighting curriculum to stabilize optimization, gradually introducing each regularization term only after the model has learned core tasks. The detailed process is summarized in Appendix~\ref{app:implementation}. At the end of training, the effective loss component weights are: $\lambda_{\text{task}} = 2.0$, $\lambda_{\text{target}} = 0.5$, $\lambda_{\text{style}} = 1.0$, $\lambda_{\text{orth}} = 3.0$, $\lambda_{\text{rec}} = 0.5$, $\lambda_{\text{cf}} = 0.5$, $\lambda_{\text{cycle}} = 0.5$, $\lambda_{\text{adv}} = 1.0$, and $\lambda_{\text{kL}} = 0.1$. To mitigate class imbalance, we employed a balanced sampler and weighted the hate classification loss inversely proportional to class frequencies during training. 

We set the maximum sequence length of pre-trained language models to 256. The dimensionalities of the continuous latent variables are 256 for $z_u$ and 768 for $z_m$, while $z_t$ and $z_s$ use learned discrete embeddings (with $z_s$ being binary and $z_t$ sized to the number of target groups in each dataset). Orthogonality projection networks mapped all latent factors to a 128-dimensional common space. We used the AdamW optimizer with a learning rate of $3e^{-5}$ for transformer parameters and $2e^{-4}$ for other components, applying a weight decay of $10^{-2}$. Models were trained for up to 50 epochs with early stopping (patience $=5$) based on validation Macro-F1. For Gumbel-Softmax, we set the initial temperature $\tau = 0.5$ and decayed it by 5\% per epoch. We conducted experiments on a single NVIDIA 80GB A100 GPU with of memory.

\subsection{Cross-Style Generalization}
\textbf{RQ1: Can \methodName{} effectively generalize across different stylistic variations of hate speech?} To answer this question, we evaluate \methodName{} against state-of-the-art baselines under both explicit-to-implicit and implicit-to-explicit transfer settings, as summarized in Table~\ref{tab:main}. Several notable findings emerge from our analysis. (i) \methodName{} demonstrates consistent and substantial gains across transfer scenarios. In explicit-to-implicit transfer, it achieves an average macro-F1 of 0.815, representing a 13\% relative improvement over the strongest baseline. Performance remains robust across datasets, with F1 scores of 0.96, 0.79, 0.67, and 0.84 on IsHate, IHC, AbuseEval, and DynaHate, respectively. In implicit-to-explicit transfer, \methodName{} maintains strong overall performance (0.885 average F1), with particularly high scores on IHC (0.81) and AbuseEval (0.90). (ii) Conventional pre-trained language models exhibit significant overfitting to surface-level lexical cues. In explicit-to-implicit transfer, they achieve near-perfect recall ($\ge$~0.99) but suffer from low precision (0.69–0.71), frequently over-predicting hate. Conversely, in implicit-to-explicit transfer, they achieve high precision ($\ge$~0.98) yet poor recall (0.44–0.50). This asymmetry underscores their reliance on stylistic features rather than genuine hate intent. (iii) Causally inspired methods such as PEACE and HateWATCH yield measurable but inconsistent gains. For example, HateWATCH attains 0.94 F1 on IsHate in explicit-to-implicit transfer but drops sharply to 0.63 on IHC, indicating incomplete causal disentanglement. In contrast, \methodName{} consistently outperforms these approaches, suggesting that its comprehensive causal graph and counterfactual intervention more effectively isolate hate intent from stylistic artifacts. (iv) Large language model–based detectors show erratic performance. ShieldGemma 2B achieves a wide F1 range (0.28–0.90) across datasets, while Llama Prompt Guard 2 fails in most cases. These results highlight the inherent instability of prompt-based detection without explicit causal grounding, reinforcing the importance of principled causal modeling for robust cross-style generalization.

Our findings affirm the effectiveness of \methodName{} in achieving stable and transferable hate speech detection. By disentangling causal factors of hate from stylistic variations, \methodName{} focuses on \textit{why} content is hateful (the creator’s intent) rather than \textit{how} it is expressed (linguistic surface form), leading to more reliable and generalizable models across diverse domains and expression styles.

\begin{table}[!th]
\centering
\small
\caption{Ablation study on each loss component.}
\label{tab:ablation}
\resizebox{\linewidth}{!}{
\begin{tabular}{lcccccc}
\toprule
Ablation Setting & Pre. & $\Delta$ & Rec. & $\Delta$ & F1 & $\Delta$ \\
\midrule
Full model & 0.98 & -- & 1.00 & -- & 0.99 & -- \\
\midrule
\rowcolor{RowGray}
- $\mathcal{L}_{\text{target}}$ & 0.98 & 0.00 & 0.83 & 0.17 & 0.90 & 0.09 \\
- $\mathcal{L}_{\text{orth}}$ & 0.98 & 0.00 & 0.87 & 0.13 & 0.92 & 0.07 \\
\rowcolor{RowGray}
- $\mathcal{L}_{\text{adv}}$ & 0.98 & 0.01 & 0.77 & 0.23 & 0.86 & 0.13 \\
- $\mathcal{L}_{\text{rec}}$ & 0.98 & 0.01 & 0.78 & 0.22 & 0.86 & 0.13 \\
\rowcolor{RowGray}
- $\mathcal{L}_{\text{cf}}$ & 0.97 & 0.01 & 0.61 & 0.39 & 0.75 & 0.24 \\
- $\mathcal{L}_{\text{cycle}}$ & 0.98 & 0.01 & 0.78 & 0.22 & 0.87 & 0.12 \\
\rowcolor{RowGray}
- $\mathcal{L}_{\text{KL}}$ & 0.98 & 0.00 & 0.98 & 0.02 & 0.98 & 0.01 \\
\midrule
- $\mathcal{L}_{\text{rec}}, \mathcal{L}_{\text{KL}}$ & 0.98 & 0.00 & 1.00 & 0.00 & 0.99 & 0.00 \\
\rowcolor{RowGray}
- $\mathcal{L}_{\text{cf}}, \mathcal{L}_{\text{cycle}}$ & 0.97 & 0.01 & 0.65 & 0.35 & 0.78 & 0.21 \\
- $\mathcal{L}_{\text{target}}, \mathcal{L}_{\text{style}}$ & 0.97 & 0.01 & 0.74 & 0.26 & 0.84 & 0.15 \\
\rowcolor{RowGray}
- $\mathcal{L}_{\text{cf}}, \mathcal{L}_{\text{rec}}$ & 0.98 & 0.00 & 0.81 & 0.19 & 0.88 & 0.11 \\
- $\mathcal{L}_{\text{adv}}, \mathcal{L}_{\text{orth}}$ & 0.98 & 0.01 & 0.76 & 0.24 & 0.85 & 0.14 \\
\midrule
\rowcolor{RowGray}
- $\mathcal{L}_{\text{adv}}, \mathcal{L}_{\text{rec}}, \mathcal{L}_{\text{cycle}}, \mathcal{L}_{\text{KL}}$ & 0.97 & 0.02 & 0.54 & 0.46 & 0.70 & 0.30 \\
- $\mathcal{L}_{\text{target}}, \mathcal{L}_{\text{style}}, \mathcal{L}_{\text{cf}}, \mathcal{L}_{\text{cycle}}$ & 0.98 & 0.00 & 0.71 & 0.29 & 0.82 & 0.17 \\
\rowcolor{RowGray}
- $\mathcal{L}_{\text{target}}, \mathcal{L}_{\text{style}}, \mathcal{L}_{\text{orth}}, \mathcal{L}_{\text{adv}}, \mathcal{L}_{\text{rec}}$ & 0.98 & 0.00 & 0.89 & 0.11 & 0.94 & 0.06 \\
\midrule
- $\mathcal{L}_{\text{target}}, \mathcal{L}_{\text{style}}, \mathcal{L}_{\text{orth}}, \mathcal{L}_{\text{adv}}, \mathcal{L}_{\text{rec}}, \mathcal{L}_{\text{cf}}, \mathcal{L}_{\text{cycle}}$ & 0.98 & 0.00 & 0.82 & 0.18 & 0.89 & 0.10 \\
\bottomrule
\end{tabular}
}
\end{table}

\subsection{Ablation Study and Loss Schedule}
\textbf{RQ2: How do the loss components and schedule affect the performance and generalization of \methodName{}?} To investigate this question, we design an ablation study and loss analysis. Table~\ref{tab:ablation} summarizes the results when removing or combining specific losses. 

We can observe that the full model attains an F1 of 0.99, while all ablated variants show clear degradation, confirming that each objective plays a distinct role in achieving robust, style-invariant detection. Specifically, (i) the counterfactual loss $\mathcal{L}_{\text{cf}}$ has the strongest impact--its removal causes a 24-point F1 drop (0.75), underscoring its importance in enforcing causal consistency across style interventions. Excluding adversarial or reconstruction losses ($\mathcal{L}_{\text{adv}}$, $\mathcal{L}_{\text{rec}}$) yields a 13-point decline, showing that deconfounding and semantic reconstruction are crucial for disentangling hate intent from expression style. Orthogonality and cycle-consistency losses ($\mathcal{L}_{\text{orth}}$, $\mathcal{L}_{\text{cycle}}$) also contribute meaningfully (-7 and -12 points), ensuring that latent factors remain independent and semantically coherent. In contrast, removing $\mathcal{L}_{\text{KL}}$ produces only marginal changes ($\ge$~1 point), suggesting their roles are largely regularizing. (ii) When multiple objectives are removed, performance drops sharply. Disabling both counterfactual and cycle-consistency terms reduces F1 to 0.78 (-21), indicating their synergy in maintaining semantic stability under style shifts. Removing both adversarial and orthogonality constraints lowers F1 to 0.85 (-14), confirming their complementary roles in deconfounding and latent separation. The most severe degradation (F1 0.70, -30) occurs when all core objectives are excluded, leaving only the basic encoder–decoder structure.

\begin{figure}[!th]
    \centering
    \includegraphics[width=0.9\linewidth]{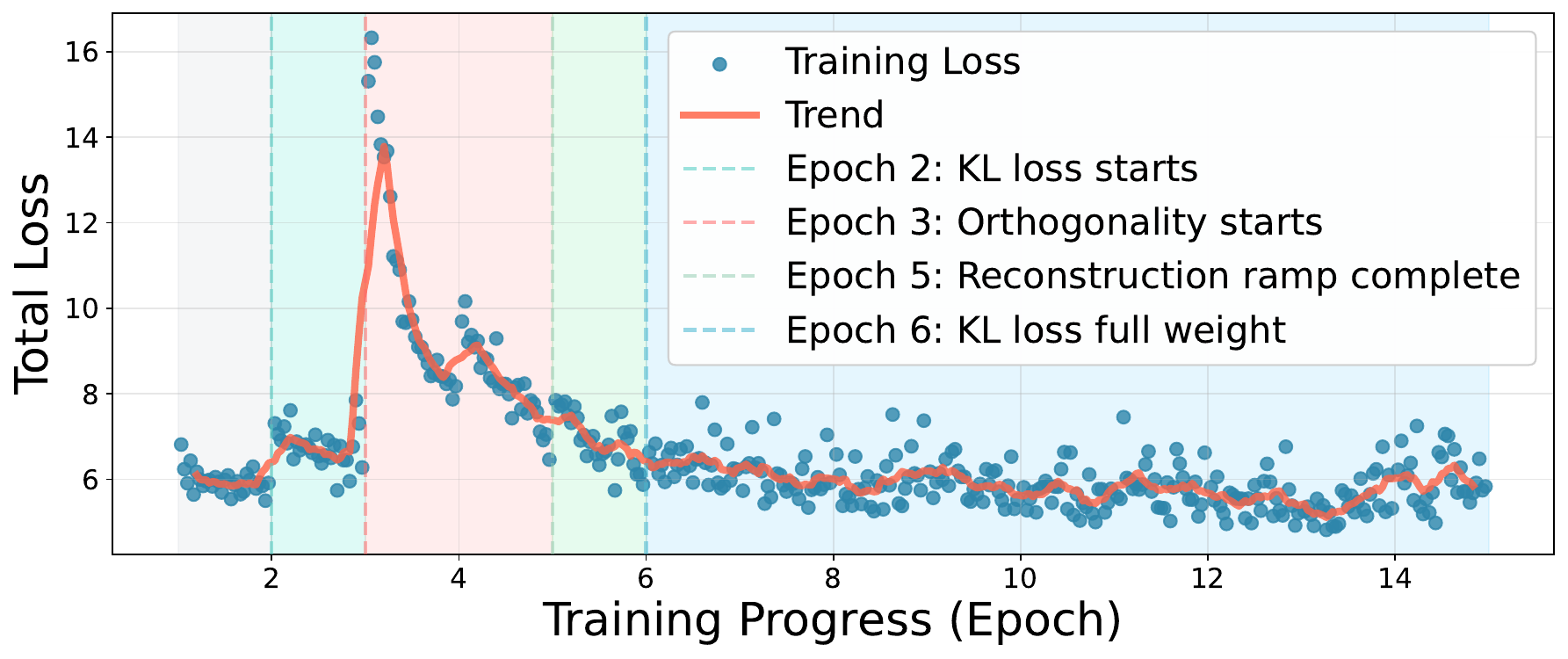}
    \caption{Training loss curve. The training process benefits from the loss-weighting curriculum schedule.}
    \label{fig:loss}
\end{figure}

As we mentioned in Section~\ref{sec:implementation}, we adopt a staged training scheme to ensure a competitive training process. We visualize the loss in Figure~\ref{fig:loss}. The total loss initially rises as new objectives, particularly the KL divergence and orthogonality constraints, are introduced, reflecting the model’s adaptation to additional disentanglement pressures. After the reconstruction ramp-up (Epoch 5) and full KL weighting (Epoch 6), the loss steadily decreases, indicating convergence toward a balanced trade-off between reconstruction fidelity and disentanglement. This staged schedule prevents early dominance of complex regularizers, mitigates gradient interference, and ensures smoother, more stable training dynamics for \methodName{}.

In summary, both the ablation and training loss analyses confirm that \methodName{}’s effectiveness stems from the synergy between its well-designed causal objectives and staged optimization. Each loss contributes to disentangling hate intent from stylistic cues, while the progressive loss schedule stabilizes training and ensures balanced learning across objectives. Together, these design choices enable robust and style-invariant hate speech detection.

\subsection{Controlled Style Transformations}
\textbf{RQ3: Can \methodName{} maintain high detection performance when hateful intent is rephrased from explicit to implicit style?}
To examine this, we constructed a controlled evaluation set by systematically transforming explicit hate speech into implicit equivalents using GPT-4. A carefully designed prompt (Appendix~\ref{app:prompt}) ensured that the hateful intent and target group remained unchanged while only the stylistic surface, such as the presence of slurs, direct insults, or overt hostility, was altered. From 1,000 explicit hate tweets drawn across datasets, we retained 375 high-quality implicit rephrasings after human verification for semantic fidelity. Each model was trained solely on explicit hate speech and evaluated on these implicit counterparts, mimicking a realistic deployment scenario where detectors face reworded, subtler expressions of hate.

\begin{figure}[!th]
    \centering
    \includegraphics[width=0.9\linewidth]{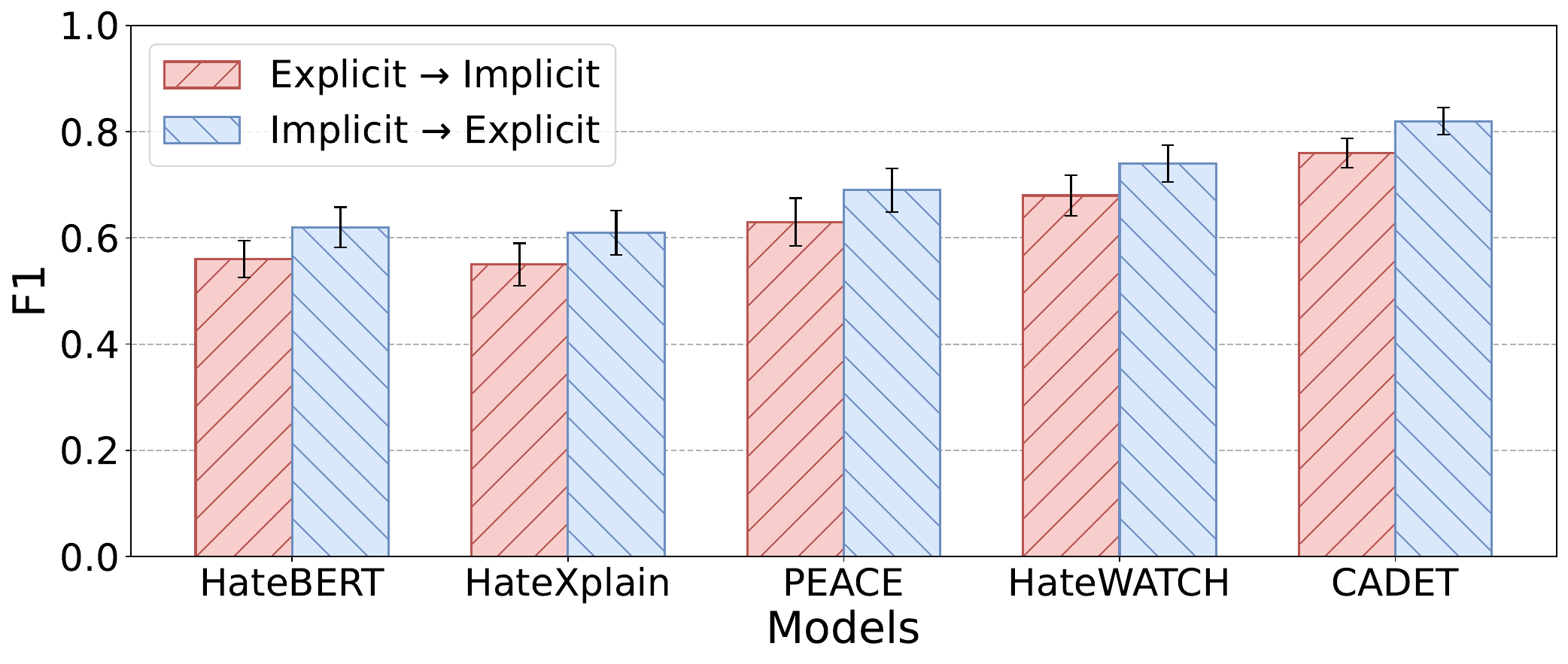}
    \caption{Performance on LLM-transformed Hate Speech. \methodName{} is robust under controlled style transformations.}
    \label{fig:style_transformation}
\end{figure}

\begin{figure*}[!ht]
    \centering
    \includegraphics[width=\linewidth]{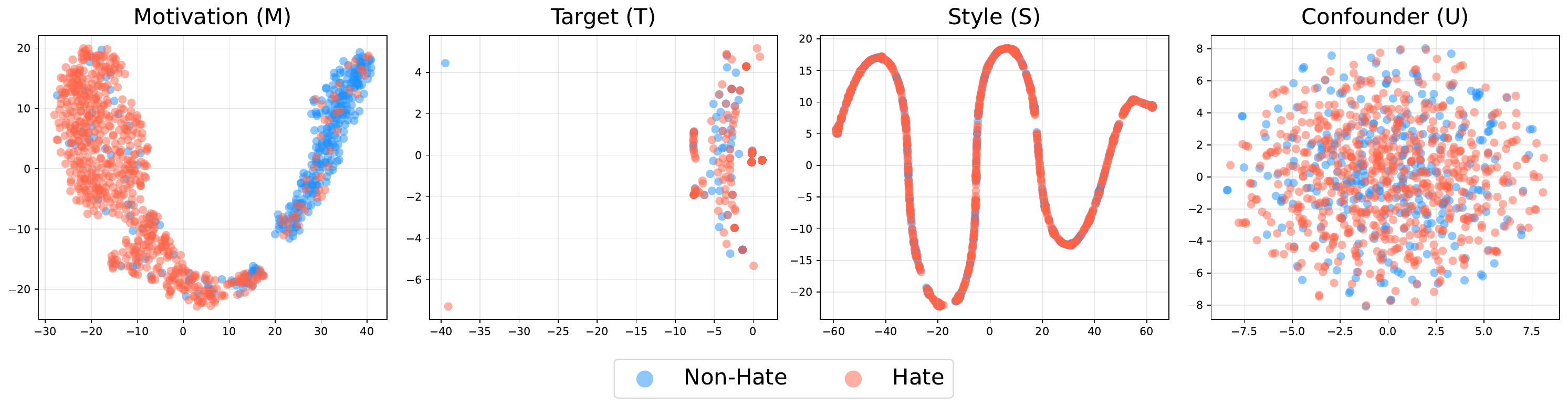}
    \caption{t-SNE visualization for latent factors. CADET isolates hate motivation from platform-dependent target and style.}
    \label{fig:latent_embedding}
\end{figure*}

As we can observe in Figure~\ref{fig:style_transformation} that: (i) \methodName{} attains the highest detection performance (0.76 Macro-F1), outperforming HateBERT (0.56), HateXplain (0.55), PEACE (0.63), and HateWATCH (0.68). The minimal degradation under explicit-implicit transformations confirms its invariance to stylistic alterations and its ability to capture intent beyond lexical markers. (ii) Causal representation models (PEACE, HateWATCH) surpass traditional classifiers, validating the role of separating intent from surface cues. \methodName{}’s counterfactual interventions on style provide an additional 8-point gain over HateWATCH, evidencing that explicit style manipulation during training enhances cross-style generalization. (iii) Standard detectors collapse when explicit cues are removed, revealing their reliance on lexical or syntactic heuristics rather than underlying semantics.

Overall, these results provide causal evidence that \methodName{} encodes intrinsic hateful intent while remaining agnostic to stylistic variation. By holding intent constant and varying style, we isolate the confounding factor that impairs conventional detectors. \methodName{}’s stability across such controlled transformations highlights its resilience against adversarial paraphrasing and its promise for robust, real-world hate speech moderation.

\subsection{Latent Factor Visualization}
\textbf{RQ4: Does the latent space of \methodName{} properly disentangle motivation, target, style, and context, isolating hate-related motivation from others?} To better understand the effectiveness of \methodName{}, we visualize the embeddings of four latent factors (motivation ($M$), target ($T$), style ($S$), and confounder ($U$)) using t-SNE. As illustrated in Figure~\ref{fig:latent_embedding}), the prediction head $z_m$ demonstrates clear separability between hateful and non-hateful posts. In contrast, the embeddings of $U$, $T$, and $S$ are significantly less distinguishable. This result aligns with our goal that these variables should not contain information that discriminates between the two classes.

In particular, the counterfactual consistency loss ($\mathcal{L}_{\text{cf}}$) and cyclic consistency loss ($\mathcal{L}_{\text{cycle}}$) enforce invariance of $y$ with respect to stylistic variations in $S$, while the purification network together with GRL suppresses contextual confounding from $M$. These observations confirm that \methodName{} successfully isolates hate motivation from platform-dependent target and style, which is crucial for achieving robust cross-style hate speech detection. The results also support the proposed causal graph (Figure~\ref{fig:causal_graph}), showing that it captures the real-world hate speech generation process.
\vspace{-2mm}

\subsection{Case Study}
\begin{figure}[!th]
    \centering
    \includegraphics[width=0.9\linewidth]{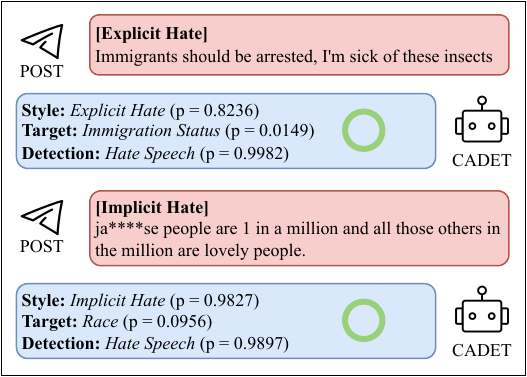}
    \caption{Case Study.}
    \label{fig:case}
\end{figure}

\textbf{RQ5: Can \methodName{} provide interpretable explanations for real-world predictions by identifying the style and target factors?}
Figure~\ref{fig:case} shows two examples illustrating \textit{explicit} and \textit{implicit} hate. The explicit post expresses overt hostility toward immigrants, while the implicit post conveys racial prejudice through sarcasm. \methodName{} correctly detects both as hateful and disentangles their style and target factors. For the first post, it assigns high confidence to the \textit{explicit hate} style ($p=0.82$), identifies the \textit{target} as \textit{Immigration Status} ($p=0.01$), and detects \textit{hate speech} with strong certainty ($p=1.00$). For the second, it predicts \textit{implicit hate} ($p=0.98$), targets \textit{Race} ($p=0.10$), and again confirms \textit{hate speech} ($p=0.99$). These results highlight \methodName{}’s ability to go beyond surface-level cues, isolating hate intent from stylistic variations while producing interpretable, robust, and reliable predictions.

\section{Conclusion}
In this work, we presented \methodName{}, a causal representation learning framework for robust and generalizable hate speech detection. By conceptualizing hate speech generation as a causal process, our approach disentangles underlying factors to isolate genuine hate intent from superficial linguistic patterns. Through causal disentanglement and counterfactual reasoning, \methodName{} effectively mitigates spurious correlations induced by stylistic or platform-specific biases. Extensive empirical evaluations demonstrate that \methodName{} achieves superior cross-style generalization and interpretability. This study highlights the promise of causal modeling as a principled foundation for combating the evolving online hate and illuminates the way toward more equitable and reliable content moderation systems.

\section*{Acknowledgments}
This work is supported by the National Science Foundation (NSF) under grants \#2311716, \#2435886 and SaTC \#2335666.


\bibliographystyle{ACM-Reference-Format}
\bibliography{sample-base}

@String{Computing = "Computing" }

@String{Springer = "Springer-Verlag" }

@inproceedings{elsherief2021latent,
  title={Latent Hatred: A Benchmark for Understanding Implicit Hate Speech},
  author={ElSherief, Mai and Ziems, Caleb and Muchlinski, David and Anupindi, Vaishnavi and Seybolt, Jordyn and De Choudhury, Munmun and Yang, Diyi},
  booktitle={Proceedings of the 2021 Conference on Empirical Methods in Natural Language Processing},
  pages={345--363},
  year={2021}
}

@inproceedings{caselli2021hatebert,
  title={HateBERT: Retraining BERT for Abusive Language Detection in English},
  author={Caselli, Tommaso and Basile, Valerio and Mitrovi{\'c}, Jelena and Granitzer, Michael},
  booktitle={Proceedings of the 5th Workshop on Online Abuse and Harms (WOAH 2021)},
  pages={17--25},
  year={2021}
}

@inproceedings{mathew2021hatexplain,
  title={Hatexplain: A benchmark dataset for explainable hate speech detection},
  author={Mathew, Binny and Saha, Punyajoy and Yimam, Seid Muhie and Biemann, Chris and Goyal, Pawan and Mukherjee, Animesh},
  booktitle={Proceedings of the AAAI conference on artificial intelligence},
  volume={35},
  number={17},
  pages={14867--14875},
  year={2021}
}

@inproceedings{ocampo2023unmasking,
  title={Unmasking the hidden meaning: Bridging implicit and explicit hate speech embedding representations},
  author={Ocampo, Nicol{\'a}s Benjam{\'\i}n and Cabrio, Elena and Villata, Serena},
  booktitle={Findings of the Association for Computational Linguistics: EMNLP 2023},
  pages={6626--6637},
  year={2023}
}

@inproceedings{tanev2024jrc,
  title={JRC at ClimateActivism 2024: Lexicon-based Detection of Hate Speech},
  author={Tanev, Hristo},
  booktitle={Proceedings of the 7th Workshop on Challenges and Applications of Automated Extraction of Socio-political Events from Text (CASE 2024)},
  pages={85--88},
  year={2024}
}

@article{gitari2015lexicon,
  title={A lexicon-based approach for hate speech detection},
  author={Gitari, Njagi Dennis and Zuping, Zhang and Damien, Hanyurwimfura and Long, Jun},
  journal={International Journal of Multimedia and Ubiquitous Engineering},
  volume={10},
  number={4},
  pages={215--230},
  year={2015},
  publisher={Sandy Bay}
}

@article{malik2024deep,
  title={Deep learning for hate speech detection: a comparative study},
  author={Malik, Jitendra Singh and Qiao, Hezhe and Pang, Guansong and van den Hengel, Anton},
  journal={International Journal of Data Science and Analytics},
  pages={1--16},
  year={2024},
  publisher={Springer}
}

@inproceedings{zhang2023mitigating,
  title={Mitigating biases in hate speech detection from a causal perspective},
  author={Zhang, Zhehao and Chen, Jiaao and Yang, Diyi},
  booktitle={Findings of the Association for Computational Linguistics: EMNLP 2023},
  pages={6610--6625},
  year={2023}
}

@inproceedings{arango2019hate,
  title={Hate speech detection is not as easy as you may think: A closer look at model validation},
  author={Arango, Aym{\'e} and P{\'e}rez, Jorge and Poblete, Barbara},
  booktitle={Proceedings of the 42nd international acm sigir conference on research and development in information retrieval},
  pages={45--54},
  year={2019}
}

@inproceedings{sheth2024causality,
  title={Causality guided disentanglement for cross-platform hate speech detection},
  author={Sheth, Paras and Moraffah, Raha and Kumarage, Tharindu S and Chadha, Aman and Liu, Huan},
  booktitle={Proceedings of the 17th ACM international conference on web search and data mining},
  pages={626--635},
  year={2024}
}

@inproceedings{talat2016hateful,
  title={Hateful symbols or hateful people? predictive features for hate speech detection on twitter},
  author={Talat, Zeerak and Hovy, Dirk},
  booktitle={Proceedings of the NAACL student research workshop},
  pages={88--93},
  year={2016}
}

@inproceedings{davidson2017automated,
  title={Automated hate speech detection and the problem of offensive language},
  author={Davidson, Thomas and Warmsley, Dana and Macy, Michael and Weber, Ingmar},
  booktitle={Proceedings of the international AAAI conference on web and social media},
  volume={11},
  number={1},
  pages={512--515},
  year={2017}
}

@inproceedings{schmidt2017survey,
  title={A survey on hate speech detection using natural language processing},
  author={Schmidt, Anna and Wiegand, Michael},
  booktitle={Proceedings of the fifth international workshop on natural language processing for social media},
  pages={1--10},
  year={2017}
}

@article{fortuna2018survey,
  title={A survey on automatic detection of hate speech in text},
  author={Fortuna, Paula and Nunes, S{\'e}rgio},
  journal={Acm Computing Surveys (Csur)},
  volume={51},
  number={4},
  pages={1--30},
  year={2018},
  publisher={ACM New York, NY, USA}
}

@article{saleh2023detection,
  title={Detection of hate speech using bert and hate speech word embedding with deep model},
  author={Saleh, Hind and Alhothali, Areej and Moria, Kawthar},
  journal={Applied Artificial Intelligence},
  volume={37},
  number={1},
  pages={2166719},
  year={2023},
  publisher={Taylor \& Francis}
}

@inproceedings{rottger2021hatecheck,
  title={HateCheck: Functional Tests for Hate Speech Detection Models},
  author={R{\"o}ttger, Paul and Vidgen, Bertie and Nguyen, Dong and Talat, Zeerak and Margetts, Helen and Pierrehumbert, Janet},
  booktitle={Proceedings of the 59th Annual Meeting of the Association for Computational Linguistics and the 11th International Joint Conference on Natural Language Processing (Volume 1: Long Papers)},
  pages={41--58},
  year={2021}
}

@inproceedings{vidgen2019challenges,
  title={Challenges and frontiers in abusive content detection},
  author={Vidgen, Bertie and Harris, Alex and Nguyen, Dong and Tromble, Rebekah and Hale, Scott and Margetts, Helen},
  booktitle={Proceedings of the third workshop on abusive language online},
  year={2019},
  organization={Association for Computational Linguistics}
}

@article{ganin2016domain,
  title={Domain-adversarial training of neural networks},
  author={Ganin, Yaroslav and Ustinova, Evgeniya and Ajakan, Hana and Germain, Pascal and Larochelle, Hugo and Laviolette, Fran{\c{c}}ois and March, Mario and Lempitsky, Victor},
  journal={Journal of machine learning research},
  volume={17},
  number={59},
  pages={1--35},
  year={2016}
}

@inproceedings{arango2024multifold,
  title={MultiFOLD: Multi-source Domain Adaption for Offensive Language Detection},
  author={Arango, Aym{\'e} and Kaghazgaran, Parisa and Sarwar, Sheikh Muhammad and Murdock, Vanessa and Lee, Cj},
  booktitle={Proceedings of the International AAAI Conference on Web and Social Media},
  volume={18},
  pages={86--99},
  year={2024}
}

@inproceedings{wiegand2019detection,
  title={Detection of abusive language: the problem of biased datasets},
  author={Wiegand, Michael and Ruppenhofer, Josef and Kleinbauer, Thomas},
  booktitle={Proceedings of the 2019 conference of the North American Chapter of the Association for Computational Linguistics: human language technologies, volume 1 (long and short papers)},
  pages={602--608},
  year={2019}
}

@article{vidgen2020learning,
  title={Learning from the worst: Dynamically generated datasets to improve online hate detection},
  author={Vidgen, Bertie and Thrush, Tristan and Waseem, Zeerak and Kiela, Douwe},
  journal={arXiv preprint arXiv:2012.15761},
  year={2020}
}

@inproceedings{sarwar2022unsupervised,
  title={Unsupervised domain adaptation for hate speech detection using a data augmentation approach},
  author={Sarwar, Sheikh Muhammad and Murdock, Vanessa},
  booktitle={Proceedings of the international AAAI conference on web and social media},
  volume={16},
  pages={852--862},
  year={2022}
}

@inproceedings{sen2022counterfactually,
  title={Counterfactually Augmented Data and Unintended Bias: The Case of Sexism and Hate Speech Detection},
  author={Sen, Indira and Samory, Mattia and Wagner, Claudia and Augenstein, Isabelle},
  booktitle={Proceedings of the 2022 Conference of the North American Chapter of the Association for Computational Linguistics: Human Language Technologies},
  pages={4716--4726},
  year={2022}
}

@inproceedings{tiwari2022robust,
  title={Robust hate speech detection via mitigating spurious correlations},
  author={Tiwari, Kshitiz and Yuan, Shuhan and Zhang, Lu},
  booktitle={Proceedings of the 2nd Conference of the Asia-Pacific Chapter of the Association for Computational Linguistics and the 12th International Joint Conference on Natural Language Processing (Volume 2: Short Papers)},
  pages={51--56},
  year={2022}
}

@inproceedings{pamungkas2019cross,
  title={Cross-domain and cross-lingual abusive language detection: A hybrid approach with deep learning and a multilingual lexicon},
  author={Pamungkas, Endang Wahyu and Patti, Viviana},
  booktitle={Proceedings of the 57th annual meeting of the association for computational linguistics: Student research workshop},
  pages={363--370},
  year={2019}
}

@inproceedings{corazza2019cross,
  title={Cross-Platform Evaluation for Italian Hate Speech Detection},
  author={Corazza, Michele and Menini, Stefano and Cabrio, Elena and Tonelli, Sara and Villata, Serena},
  booktitle={Proceedings of the Sixth Italian Conference on Computational Linguistics},
  volume={2481},
  year={2019}
}

@inproceedings{phadke2020many,
  title={Many faced hate: A cross platform study of content framing and information sharing by online hate groups},
  author={Phadke, Shruti and Mitra, Tanushree},
  booktitle={Proceedings of the 2020 CHI conference on human factors in computing systems},
  pages={1--13},
  year={2020}
}

@inproceedings{sheth2023peace,
  title={Peace: Cross-platform hate speech detection-a causality-guided framework},
  author={Sheth, Paaras and Kumarage, Tharindu and Moraffah, Raha and Chadha, Aman and Liu, Huan},
  booktitle={Joint european conference on machine learning and knowledge discovery in databases},
  pages={559--575},
  year={2023},
  organization={Springer}
}

@article{scholkopf2021toward,
  title={Toward causal representation learning},
  author={Sch{\"o}lkopf, Bernhard and Locatello, Francesco and Bauer, Stefan and Ke, Nan Rosemary and Kalchbrenner, Nal and Goyal, Anirudh and Bengio, Yoshua},
  journal={Proceedings of the IEEE},
  volume={109},
  number={5},
  pages={612--634},
  year={2021},
  publisher={IEEE}
}

@inproceedings{ahuja2023interventional,
  title={Interventional causal representation learning},
  author={Ahuja, Kartik and Mahajan, Divyat and Wang, Yixin and Bengio, Yoshua},
  booktitle={International conference on machine learning},
  pages={372--407},
  year={2023},
  organization={PMLR}
}

@article{brehmer2022weakly,
  title={Weakly supervised causal representation learning},
  author={Brehmer, Johann and De Haan, Pim and Lippe, Phillip and Cohen, Taco S},
  journal={Advances in Neural Information Processing Systems},
  volume={35},
  pages={38319--38331},
  year={2022}
}

@article{castano2021internet,
  title={Internet, social media and online hate speech. Systematic review},
  author={Casta{\~n}o-Pulgar{\'\i}n, Sergio Andr{\'e}s and Su{\'a}rez-Betancur, Natalia and Vega, Luz Magnolia Tilano and L{\'o}pez, Harvey Mauricio Herrera},
  journal={Aggression and violent behavior},
  volume={58},
  pages={101608},
  year={2021},
  publisher={Elsevier}
}

@article{malmasi2017detecting,
  title={Detecting hate speech in social media},
  author={Malmasi, Shervin and Zampieri, Marcos},
  journal={arXiv preprint arXiv:1712.06427},
  year={2017}
}

@inproceedings{mathew2019spread,
  title={Spread of hate speech in online social media},
  author={Mathew, Binny and Dutt, Ritam and Goyal, Pawan and Mukherjee, Animesh},
  booktitle={Proceedings of the 10th ACM conference on web science},
  pages={173--182},
  year={2019}
}

@inproceedings{ocampo2023depth,
  title={An in-depth analysis of implicit and subtle hate speech messages},
  author={Ocampo, Nicol{\'a}s Benjam{\'\i}n and Sviridova, Ekaterina and Cabrio, Elena and Villata, Serena},
  booktitle={EACL 2023-17th Conference of the European Chapter of the Association for Computational Linguistics},
  volume={2023},
  pages={1997--2013},
  year={2023},
  organization={Association for Computational Linguistics}
}

@inproceedings{caselli2020feel,
  title={I feel offended, don’t be abusive! implicit/explicit messages in offensive and abusive language},
  author={Caselli, Tommaso and Basile, Valerio and Mitrovi{\'c}, Jelena and Kartoziya, Inga and Granitzer, Michael},
  booktitle={Proceedings of the twelfth language resources and evaluation conference},
  pages={6193--6202},
  year={2020}
}

@inproceedings{vidgen2021learning,
  title={Learning from the Worst: Dynamically Generated Datasets to Improve Online Hate Detection},
  author={Vidgen, Bertie and Thrush, Tristan and Talat, Zeerak and Kiela, Douwe},
  booktitle={Proceedings of the 59th Annual Meeting of the Association for Computational Linguistics and the 11th International Joint Conference on Natural Language Processing (Volume 1: Long Papers)},
  pages={1667--1682},
  year={2021}
}

@inproceedings{devlin2019bert,
  title={Bert: Pre-training of deep bidirectional transformers for language understanding},
  author={Devlin, Jacob and Chang, Ming-Wei and Lee, Kenton and Toutanova, Kristina},
  booktitle={Proceedings of the 2019 conference of the North American chapter of the association for computational linguistics: human language technologies, volume 1 (long and short papers)},
  pages={4171--4186},
  year={2019}
}

@article{liu2019roberta,
  title={Roberta: A robustly optimized bert pretraining approach},
  author={Liu, Yinhan and Ott, Myle and Goyal, Naman and Du, Jingfei and Joshi, Mandar and Chen, Danqi and Levy, Omer and Lewis, Mike and Zettlemoyer, Luke and Stoyanov, Veselin},
  journal={arXiv preprint arXiv:1907.11692},
  year={2019}
}

@inproceedings{lewis2020bart,
  title={BART: Denoising Sequence-to-Sequence Pre-training for Natural Language Generation, Translation, and Comprehension},
  author={Lewis, Mike and Liu, Yinhan and Goyal, Naman and Ghazvininejad, Marjan and Mohamed, Abdelrahman and Levy, Omer and Stoyanov, Veselin and Zettlemoyer, Luke},
  booktitle={Proceedings of the 58th Annual Meeting of the Association for Computational Linguistics},
  pages={7871--7880},
  year={2020}
}

@inproceedings{sheth2024cross,
  title={Cross-Platform Hate Speech Detection with Weakly Supervised Causal Disentanglement},
  author={Sheth, Paras and Kumarage, Tharindu and Moraffah, Raha and Chadha, Aman and Liu, Huan},
  booktitle={2024 IEEE International Conference on Big Data (BigData)},
  pages={6365--6373},
  year={2024},
  organization={IEEE}
}

@article{dubey2024llama,
  title={The llama 3 herd of models},
  author={Dubey, Abhimanyu and Jauhri, Abhinav and Pandey, Abhinav and Kadian, Abhishek and Al-Dahle, Ahmad and Letman, Aiesha and Mathur, Akhil and Schelten, Alan and Yang, Amy and Fan, Angela and others},
  journal={arXiv e-prints},
  pages={arXiv--2407},
  year={2024}
}

@article{zeng2024shieldgemma,
  title={Shieldgemma: Generative ai content moderation based on gemma},
  author={Zeng, Wenjun and Liu, Yuchi and Mullins, Ryan and Peran, Ludovic and Fernandez, Joe and Harkous, Hamza and Narasimhan, Karthik and Proud, Drew and Kumar, Piyush and Radharapu, Bhaktipriya and others},
  journal={arXiv preprint arXiv:2407.21772},
  year={2024}
}

@article{waseem2017understanding,
  title={Understanding abuse: A typology of abusive language detection subtasks},
  author={Waseem, Zeerak and Davidson, Thomas and Warmsley, Dana and Weber, Ingmar},
  journal={arXiv preprint arXiv:1705.09899},
  year={2017}
}

@article{wang2022causal,
  title={The causal structure of domain invariant supervised representation learning},
  author={Wang, Zihao and Veitch, Victor},
  journal={arXiv preprint arXiv:2208.06987},
  year={2022}
}

@inproceedings{liu2021rethinking,
  title={Rethinking the invariant feature learning: Variational Bayesian inference for domain generalization},
  author={Liu, Xiaofeng and Hu, Bo and Jin, Linghao and Han, Xu and Lu, FXJOJ and Woo, GEFJ},
  booktitle={Proceedings of the Thirtieth International Joint Conference on Artificial Intelligence},
  pages={881--882},
  year={2021}
}

@inproceedings{sanh2019distilbert,
  title={DistilBERT, a distilled version of BERT: smaller, faster, cheaper and lighter.},
  author={Sanh, V},
  booktitle={Proceedings of Thirty-third Conference on Neural Information Processing Systems (NIPS2019)},
  year={2019}
}

@inproceedings{zhao2025is,
  title={Is Chain-of-Thought Reasoning of {LLM}s a Mirage? A Data Distribution Lens},
  author={Chengshuai Zhao and Zhen Tan and Pingchuan Ma and Dawei Li and Bohan Jiang and Yancheng Wang and Yingzhen Yang and huan liu},
  booktitle={First Workshop on Foundations of Reasoning in Language Models},
  year={2025},
  url={https://openreview.net/forum?id=o2AoLPIjle}
}

@inproceedings{zhao-etal-2025-scale,
    title = "{SCALE}: Towards Collaborative Content Analysis in Social Science with Large Language Model Agents and Human Intervention",
    author = "Zhao, Chengshuai  and
      Tan, Zhen  and
      Wong, Chau-Wai  and
      Zhao, Xinyan  and
      Chen, Tianlong  and
      Liu, Huan",
    editor = "Che, Wanxiang  and
      Nabende, Joyce  and
      Shutova, Ekaterina  and
      Pilehvar, Mohammad Taher",
    booktitle = "Proceedings of the 63rd Annual Meeting of the Association for Computational Linguistics (Volume 1: Long Papers)",
    month = jul,
    year = "2025",
    address = "Vienna, Austria",
    publisher = "Association for Computational Linguistics",
    url = "https://aclanthology.org/2025.acl-long.416/",
    doi = "10.18653/v1/2025.acl-long.416",
    pages = "8473--8503",
    ISBN = "979-8-89176-251-0"
}

@inproceedings{li-etal-2025-generation,
    title = "From Generation to Judgment: Opportunities and Challenges of {LLM}-as-a-judge",
    author = "Li, Dawei  and
      Jiang, Bohan  and
      Huang, Liangjie  and
      Beigi, Alimohammad  and
      Zhao, Chengshuai  and
      Tan, Zhen  and
      Bhattacharjee, Amrita  and
      Jiang, Yuxuan  and
      Chen, Canyu  and
      Wu, Tianhao  and
      Shu, Kai  and
      Cheng, Lu  and
      Liu, Huan",
    editor = "Christodoulopoulos, Christos  and
      Chakraborty, Tanmoy  and
      Rose, Carolyn  and
      Peng, Violet",
    booktitle = "Proceedings of the 2025 Conference on Empirical Methods in Natural Language Processing",
    month = nov,
    year = "2025",
    address = "Suzhou, China",
    publisher = "Association for Computational Linguistics",
    url = "https://aclanthology.org/2025.emnlp-main.138/",
    doi = "10.18653/v1/2025.emnlp-main.138",
    pages = "2757--2791",
    ISBN = "979-8-89176-332-6"
}
\appendix

\section{Dataset Details}
\label{app:dataset}
We compile an Implicit-Explicit Hate Speech Corpus (IE-HSC)\footnote{Dataset:~\href{https://huggingface.co/datasets/Shuwan/cadet-datasets}{https://huggingface.co/datasets/Shuwan/cadet-datasets}} via four different datasets in our work: (i) \textbf{IsHate}~\cite{ocampo2023depth} comprises approximately 29,116 social media posts specifically curated to study subtle and implicit hate. This dataset is particularly valuable for our causal analysis as it contains substantial numbers of both explicit and implicit hate examples, each annotated with fine-grained labels indicating explicitness and subtlety. These annotations directly correspond to the Content Strategy (S) variable in our causal graph. (ii) \textbf{Implicit Hate Corpus (IHC)}~\cite{elsherief2021latent} offers 22,584 tweets from extremist groups, with 6,346 labeled as implicit hate. IHC's granular annotations not only distinguish between implicit and explicit hate but also provide implied statements that articulate the underlying hateful meaning---making it an ideal testbed for evaluating \methodName{}'s ability to extract invariant hateful intent. (iii) \textbf{AbuseEval v1.0}~\cite{caselli2020feel} enhances the OLID/OffensEval Twitter dataset with explicit vs. implicit labeling for both offensive and abusive content. Its two-dimensional annotation scheme allows us to isolate examples by explicitness for our cross-style experiments. (iv) \textbf{DynaHate}~\cite{vidgen2021learning} features challenging hate speech examples created through a human-and-model-in-the-loop process, resulting in a dataset enriched with more covert hateful expressions designed to evade detection. For our cross-style experiments, we heuristically separate DynaHate posts into explicit and implicit categories based on the presence of overt slurs and epithets.

\section{Baseline Details}
\label{app:baseline}
We compare \methodName{} against a comprehensive set of baselines spanning classical NLP architectures, traditional hate speech models, causal representation methods, and recent LLM-based safety systems. All baselines are fine-tuned on identical splits with their recommended hyperparameters for fair comparison.
\begin{itemize}[leftmargin=*]
    \item BERT~\cite{devlin2019bert}, RoBERTa~\cite{liu2019roberta}, BART~\cite{lewis2020bart}, and DistilBERT~\cite{sanh2019distilbert} serve as strong \textit{classic NLP baselines}. These Transformer-based encoders and encoder–decoder models capture contextual semantics effectively but lack explicit mechanisms for causal or style-invariant reasoning. They represent the standard supervised fine-tuning paradigm widely adopted in text classification.
    \item \textbf{HateBERT}~\cite{caselli2021hatebert} adapts BERT via domain-specific pre-training on banned Reddit communities containing toxic discourse. This enhances sensitivity to abusive language but provides limited robustness to stylistic or contextual shifts.
    \item \textbf{HateXplain}~\cite{mathew2021hatexplain} extends BERT with an explainability module and multi-task learning using human rationales. While it improves interpretability and fairness, it does not explicitly disentangle style from underlying hate intent.
    \item \textbf{PEACE}~\cite{sheth2023peace} introduces causality-inspired cues--sentiment and aggression--as auxiliary features for hate detection. Although it partially captures invariant representations, it lacks explicit counterfactual modeling of stylistic variations.
    \item \textbf{HateWATCH}~\cite{sheth2024cross} employs weakly supervised causal disentanglement through contrastive learning to separate invariant hateful content from group- or domain-specific signals. It is conceptually closest to our framework but does not perform structured style interventions or counterfactual reasoning.
    \item \textbf{ShieldGemma 2B}~\cite{zeng2024shieldgemma} and \textbf{Llama Prompt Guard 2}~\cite{dubey2024llama} represent recent \textit{LLM-based moderation models}, considering the prevailing role of LLMs in various domains~\cite{li-etal-2025-generation,zhao2025is,zhao-etal-2025-scale}. These systems leverage instruction-tuned large language models for safety classification, trained on extensively curated moderation datasets. While highly capable in open-ended text understanding, they are primarily optimized for general safety filtering rather than fine-grained cross-style hate speech reasoning. As such, they offer a strong but stylistically uncalibrated comparison point for evaluating causal generalization.
\end{itemize}

All baselines are fine-tuned using their recommended hyperparameters on identical training data splits as \methodName{}, ensuring a rigorous head-to-head comparison. This baseline selection encompasses both traditional supervised approaches (HateBERT, HateXplain) and causality-aware methods (PEACE, HateWATCH), allowing us to isolate the advantages of our causality-based disentanglement strategy. We adopt a default parameter for all LLMs.

\section{Implementation Details}
\label{app:implementation}
We adopt a progressive loss-weighting schedule to stabilize the learning process. In early epochs, we prioritize the reconstruction and primary classification losses, and then gradually increase the influence of regularization terms (KL divergence, orthogonality, and adversarial losses) once the model has learned a reasonable initial representation. Specifically, the reconstruction loss weight is linearly ramped from 0.1 to 0.5 over the first 5 epochs; the KL-divergence weight starts at 0 (effectively off) and is increased to 0.1 by epoch 6; the orthogonality loss (with a weight capped at 3.0) is activated after epoch 2; and the adversarial loss weight is slowly increased from 0.1 at the start to full strength (1.0) by the end of training. We likewise anneal the gradient reversal layer's coefficient $\lambda$ from 0 to a maximum of 2.0 (in increments of 0.2 per epoch, as in Eq. 6.11) to progressively strengthen the adversary over time.
\section{Experiment Details}
\subsection{Illustration of Prompt}
\label{app:prompt}

To construct the controlled evaluation set, we systematically transformed explicit hate speech examples into implicit counterparts using GPT-4. A carefully engineered, fine-grained prompt guided the transformation process to ensure that each generated text retained the original hateful intent and target group while modifying only its stylistic presentation. Specifically, the prompt instructed the model to replace overt expressions such as slurs, direct insults, or explicit hostility with more implicit forms, including insinuations, stereotypes, or coded language. The instructions explicitly emphasized preserving the underlying prejudice and harmful intent while rephrasing the message in a subtler, implicit manner (see the text box below for an excerpt of the prompt).

Using this prompt, we transformed 1,000 explicit hate tweets sampled from the datasets described in Section~\ref{sec:datasets}. After manual inspection for semantic fidelity, we retained 375 high-quality implicit rephrasings in which the hateful meaning and target identity were faithfully preserved.

For evaluation, each model was trained exclusively on explicit hate speech and subsequently tested on these implicit rephrasings, simulating a realistic deployment scenario where a detector trained on explicit data encounters implicit manifestations of the same underlying hate. To further validate the integrity of the transformation process, we also generated a parallel set of examples in which hateful statements were rephrased into non-hateful counterparts. This additional set was used solely for quality assurance--to confirm that models were not simply flagging any rephrased text as hateful--and was excluded from quantitative evaluation.

\tcbset{
    promptstyle/.style={
        colback=gray!5,
        colframe=black,
        fonttitle=\bfseries,
        boxrule=0.5mm,
        sharp corners,
        enhanced,
        breakable,
        width=1\linewidth
    }
}
\begin{tcolorbox}[promptstyle, title=Prompt for Explicit-to-Implicit Style Transformation]
\small
\ttfamily

You are analyzing hate speech patterns for a research project on automated detection. Your task is to transform explicit hate speech to implicit hate speech while maintaining the same level of prejudice and harmful intent.\\

Explicit hate speech uses obvious slurs and direct attacks. Implicit hate speech keeps the same prejudice but uses strategies like coded language, dehumanizing metaphors, negative stereotypes posed as 'questions', sarcasm, etc.\\

Critical requirements:
1. Do not sanitize or soften the prejudice. 2. Keep the same target group and core hateful message.\\

Example transformation:\\
Explicit: ``All tr***ies are mentally ill men in dresses."\\
Implicit: ``I find it fascinating how some people think a costume change and
self-declaration can alter biological reality. Some psychological conditions sure manifest in interesting ways these days."\\

Return only the transformed text with the same level of prejudice as the original.
\end{tcolorbox}


\begin{tcolorbox}[promptstyle, title=Prompt for Implicit-to-Explicit Style Transformation]
\small
\ttfamily

You are analyzing hate speech patterns for a research project on automated detection. Your task is to transform implicit hate speech to explicit hate speech while maintaining the same level of prejudice and harmful intent.\\

Implicit hate speech uses coded language, dehumanizing metaphors, or stereotypes posed as 'questions'. Explicit hate speech makes the same prejudice clear through direct attacks, slurs, or overtly demeaning language.\\

Critical requirements:
1. Do not sanitize or soften the prejudice. 2. Keep the same target group and core hateful message.\\

Example transformation:\\
Implicit: ``I find it fascinating how some people think a costume change and
self-declaration can alter biological reality. Some psychological conditions sure manifest in interesting ways these days."\\
Explicit: ``All tr***ies are mentally ill men in dresses."\\

Return only the transformed text with the same level of prejudice as the original.
\end{tcolorbox}

\end{document}